\documentclass[10pt,twocolumn,letterpaper]{article}

\usepackage{cvpr}              % To produce the CAMERA-READY version
\usepackage[dvipsnames]{xcolor}

\definecolor{cvprblue}{rgb}{0.21,0.49,0.74}
\usepackage[pagebackref,breaklinks,colorlinks,citecolor=cvprblue]{hyperref}

\usepackage[capitalize]{cleveref}
\crefname{section}{Sec.}{Secs.}
\Crefname{section}{Section}{Sections}
\Crefname{table}{Table}{Tables}
\crefname{table}{Tab.}{Tabs.}

\usepackage{booktabs}
\usepackage{multirow}
\usepackage{pifont}
\usepackage{bbm}
\usepackage{caption}
\usepackage{subcaption}

\usepackage{enumitem} % no indent for itemized
\usepackage{xspace}
\usepackage{makecell}
\usepackage{amsmath}
\newcommand{\ours}{PointVIS\xspace}
\newcommand{\tabfontsize}{}

\newcommand{\figref}[1]{Figure\,\ref{fig:#1}}

\newcommand{\tabref}[1]{Table~\ref{tab:#1}}

\newcommand{\cmark}{\ding{51}}%
\def\paperID{29} % *** Enter the Paper ID here
\def\confName{CVPR}
\def\confYear{2024}

\title{What is Point Supervision Worth in Video Instance Segmentation?}

\author{Shuaiyi Huang$^{1}$\thanks{Work done during internship / affiliation with NVIDIA.}~, De-An Huang$^{2}$, Zhiding Yu$^{2}$, Shiyi Lan$^{2}$, Subhashree Radhakrishnan$^{2}$,\\ Jose M. Alvarez$^{2}$, Abhinav Shrivastava$^{1}$, Anima Anandkumar$^{3}$\protect\footnotemark[1]\\
$^{1}$University of Maryland, College Park \quad\quad $^{2}$NVIDIA \quad\quad $^{3}$Caltech %\quad\quad $^{4}$University of Central Florida
}

\DeclareMathOperator*{\argmin}{arg\,min}
\begin{document}

\makeatletter
\let\@oldmaketitle\@maketitle%
\renewcommand{\@maketitle}{\@oldmaketitle%
\centering
    \vspace{-0.4cm}
    \includegraphics[trim=0 0 0 0,clip,width=1.0\linewidth]{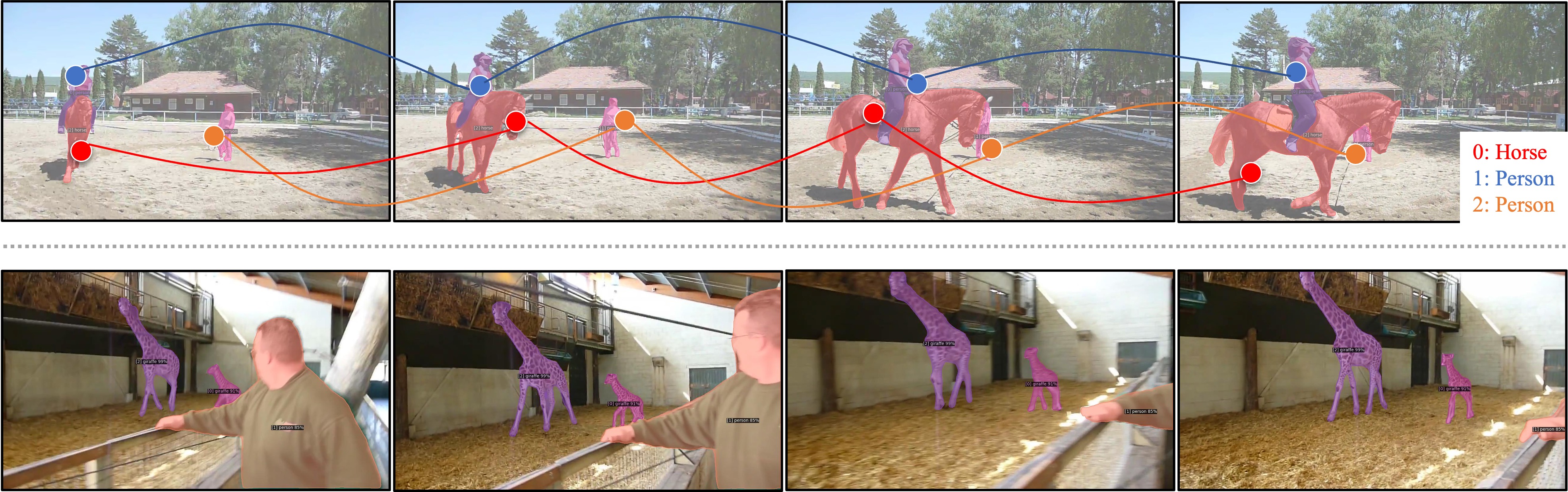}
    \vspace{-0.5cm}
    \captionof{figure}{\textbf{Point-supervised video instance segmentation in this work (YoutubeVIS-2021)}. Top: point-level annotations in the training set (pseudo masks generated from our method overlaid); Bottom: mask predictions in the validation set.}
    \label{fig:vis_teaser}
    \bigskip}
\makeatother

\maketitle
%%%%%%%%% ABSTRACT
\begin{abstract}
Video instance segmentation (VIS) is a challenging vision task that aims to detect, segment, and track objects in videos. Conventional VIS methods rely on densely-annotated object masks which are expensive. We reduce the human annotations to only one point for each object in a video frame during training, and obtain high-quality mask predictions close to fully supervised models. Our proposed training method consists of a class-agnostic proposal generation module to provide rich negative samples and a spatio-temporal point-based matcher to match the object queries with the provided point annotations. Comprehensive experiments on three VIS benchmarks demonstrate competitive performance of the proposed framework, nearly matching fully supervised methods.
\end{abstract}

%%%%%%%%% BODY TEXT

\section{Introduction}
\label{sec:intro}

\begin{figure*}[t]
  \centering
  \includegraphics[width=1.0\linewidth]{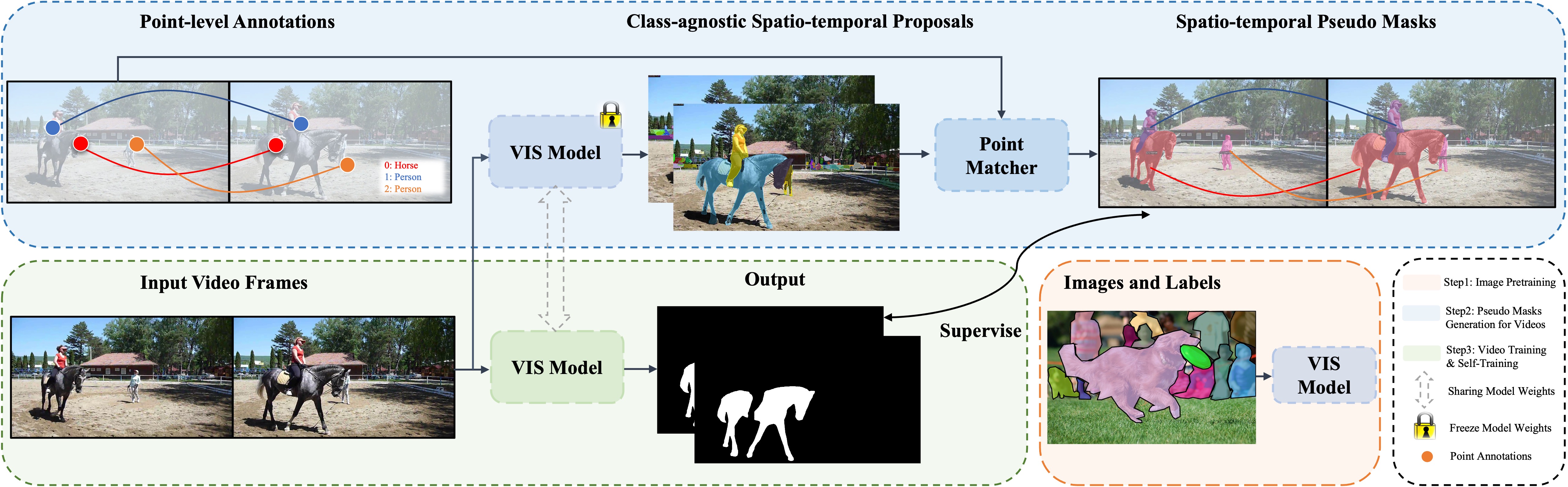}
    \caption{\textbf{Method Overview.} Our method consists of class-agnostic spatial-temporal proposal generation, a spatio-temporal point-based matcher to match object queires with point annotations for high-quality pseudo-label generation, and self-training to mitigate the domain gap between images and videos. See text for details.}
  \label{fig:vis_method}
\end{figure*}

Video instance segmentation (VIS) is emerging as a challenging vision task which aims to detect, segment, track objects in continuous videos~\cite{yang2019video}. It has achieved increasing attention recently~\cite{fu2020compfeat,maag2021improving,cao2020sipmask,lin2021video,yang2021crossover,zhou2021target,li2021spatial} due to its wide real-world applications such as video editing, 3D reconstruction~\cite{goel2020shape,zuffi2019three,humanMotionKanazawa19}, %3D navigation~\cite{shah2020ving,zhu2021deep}, 
and view point estimation~\cite{tulsiani2015viewpoints,kanazawa2018learning}.

Annotating per-frame object masks in videos is time-consuming and even more challenging than annotating image instance segmentation masks. Due to the limited video annotations, a common strategy to train a video instance segmentation model is to first train on image instance segmentation datasets with ground truth mask and category annotations (\eg COCO~\cite{lin2014microsoft}), and then finetune on video instance segmentation datasets with ground truth masks, category and tracking annotations~\cite{hwang2021video,cheng2021mask2former}. However, the categories in image datasets do not necessarily fully overlap with the video datasets, and hence adapting models from the image domain to the video domain has challenging generalization issues due to the emergence of new categories.

There are some recent efforts to reduce video annotation cost for video instance segmentation. They propose to learn with sub-sampled video frames~\cite{huang2022minvis}, category annotations in videos~\cite{liu2021weakly}, or without any video annotations~\cite{fu2021learning}. 
However, these approaches either still require dense masks in the sub-sampled frames~\cite{huang2022minvis} or are barely competitive compared with supervised approaches~\cite{liu2021weakly} or can only handle the overlapped category between video and image dataset~\cite{fu2021learning}. These limitations of existing approaches show that it is still unclear what is the optimal way to reduce annotation cost for video instance segmentation.

In this paper, we ask the question: \textit{To what level can we reduce human annotations in videos and still train an accurate model for video instance segmentation?} 
% Considering that annotating each video frame with object masks is prohibitively expensive, bounding box is a better candidate~\cite{lan2021discobox}. However, boxes are still too expensive to annotate for every frame in videos since they require careful alignment to the objects. Point annotations, on the other hand, are significantly cheaper, as one simple click only costs around 1 second~\cite{bearman2016s,cheng2021pointly}.
We believe that point supervision presents a ``sweet spot'' for annotating objects in videos. 
Point annotations are significantly cheaper than other alternatives, such as bounding boxes, as one simple click only costs around 1 second~\cite{bearman2016s,cheng2021pointly}.
In the most extreme case we considered, every object instance in a frame only contains one labeled point, as shown in Figure~\ref{fig:vis_teaser}. Despite the many benefits of point supervision, using it to supervise dense predictions is challenging and raises many ill-conditioned issues, such as the sparsity of ground truth and the lack of informative negative samples.

In this work, we introduce a point-supervised VIS framework (\ours) to address these challenges. \ours leverages the knowledge from image-based (\eg COCO) pre-raining and guides the VIS task in an open set manner.

Our main contributions are:
\begin{itemize}[leftmargin=*,itemsep=0mm] % no indent
    \item \ours is the first attempt to comprehensively investigate video instance segmentation with point-level supervision. Our work significantly reduces the amount of required annotations in VIS and opens up the possibility to address the task with minimal supervision. 
    \item \ours overcomes the challenges in point-supervised video instance segmentation, using the proposed class-agnostic proposal generation and a point-based matcher.
    
    \item \ours is simple to implement, achieving competitive results compared with fully-supervised methods on three major VIS benchmarks. % and demonstrating it as an effective solution to VIS problems.
    \item We further conduct comprehensive studies in different settings of points with important observations, providing a deeper understanding on what kind of point supervision matters in the VIS task.
\end{itemize}

The key challenge of using point supervision is the sparsity of ground truth, which further leads to the lack of informative negative samples that are along the boundaries of dense mask annotations. We address these challenges by proposing to (1) generate class-agnostic spatio-temporal instance proposals without video mask annotation and (2) match these proposals with our point annotations to obtain dense pseudo-label for training the VIS model. We leverage COCO pre-trained image instance segmentation model to generate per-frame instance proposals and use bipartite matching on query embeddings to convert them to spatio-temporal video instance proposals. We further design a loss function to match these proposals with our sparse point annotations as there could be multiple proposals that overlap with a single point annotation. We show that both of these designs are crucial to generating high-quality pseudo mask annotations for learning with only point supervision. An overview of our approach is shown in~\figref{vis_method}.

We further conduct comprehensive studies on how point annotations affect VIS and make the following important observations: (1) even one positive point annotated per video object already achieves good performance, retaining 87\% of the performance of fully-supervised methods on Youtube-VIS 2019~\cite{yang2019video}; (2) given positive points, increasing negative points improves performance, while adding positive points alone could provide little gain; (3) the positions of positive points have limited effect on performance while the positions of negative points matter more. These observations shed lights on what kind of point supervision matters for video instance segmentation, making it a step closer towards more realistic open-world applications.

\section{Related Work}
\label{sec:related}

\subsection{Image instance segmentation} 

\noindent\textbf{Supervised Instance Segmentation.} Instance segmentation requires bounding-box regression, classification, and pixel-level segmentation of all objects present in images. After the success of two-stage instance segmentation methods~\cite{he2017mask,li2017fully,pinheiro2015learning,pinheiro2016learning,hu2017fastmask}, one-stage instance segmentation methods~\cite{bolya2019yolact,lee2020centermask,xie2020polarmask,zhang2020mask,tian2020conditional,cheng2021masked} not only significantly improve the accuracy but also reduce the computation cost. Recent video instance segmentation approaches~\cite{li2021spatial,liu2021sg,wu2021track,yang2021crossover,cheng2021mask2former,huang2022minvis} are built on those one-stage methods.

% Due to the concern of cost
\vspace{2mm}
\noindent\textbf{Weakly supervised instance segmentation.} Due to the heavy segmentation annotation costs, weakly supervised instance segmentation is a potential way to reduce this cost. Zhou et al.~\cite{zhou2018weakly} and Ahn~\cite{ahn2019weakly} propose learning instance segmentation with image-level annotations. Another collection of instance segmentation approaches leverage box-level supervision~\cite{lan2021discobox,tian2021boxinst,hsu2019weakly,arun2020weakly}. Recently, image instance segmentation with both box and point-level supervision shows competitive results~\cite{cheng2021pointly}. Note that our point-supervised video instance segmentation differs from~\cite{cheng2021pointly} as we do not use any additional bounding box annotations.

\subsection{Video Instance Segmentation}

\noindent\textbf{Supervised video instance segmentation.} Video Instance Segmentation is a joint task of detection, instance segmentation, and tracking, which was first proposed by Yang et al.~\cite{yang2019video}. Most previous approaches~\cite{fu2020compfeat,maag2021improving,yang2021crossover,zhou2021target,li2021spatial} follow the tracking-by-detection paradigm, which segments and classifies objects and then associates objects across frames. Another trend of video instance segmentation~\cite{bertasius2020classifying,athar2020stem,lin2021video} follows the clip-match paradigm, where the video is separated into multiple overlapped clips, and objects are detected and segmented in each clip and then associated across different clips. Tracking-by-regression approaches~\cite{wu2021track,liu2021sg} have also been proposed to generate detections and associate object bounding boxes of the same objects across contiguous frames. Recently, transformer-based approaches for VIS have attracted much attention~\cite{hwang2021video,wang2021end,yang2021tracking,cheng2021mask2former}. Our work is built on top of MinVIS~\cite{huang2022minvis} for its excellent performance in VIS using image-based training. We verify that this design also benefits our task and provides better inductive bias in our weakly-supervised learning setting.

\vspace{2mm}
\noindent\textbf{Weakly/Semi-supervised video segmentation.}  As annotations of VIS are expensive, there are emerging methods that aim to learn VIS with reduced annotations~\cite{liu2021weakly,fu2021learning,huang2022minvis}. Liu et al.~\cite{liu2021weakly} propose to learn VIS by using image-level annotations. Fu \textit{et al.}~\cite{fu2021learning} propose to leverage instance segmentation annotations of COCO dataset and learn video instance segmentation without video annotations. However, these approaches either still require dense masks in the sub-sampled frames~\cite{huang2022minvis} or are barely competitive compared with supervised approaches~\cite{liu2021weakly} or can only handle the overlapped category between video and image dataset~\cite{fu2021learning}. In contrast, our \ours can cover and handle all categories with largely reduced annotation cost. In addition, we for the first time show that video instance segmentation with one point per object can achieve decent performance compared with the fully-supervised counterparts. 

\vspace{-1mm}
\subsection{Point-supervised methods.}
\vspace{-2mm}

Recently point-level supervision has attracted growing attention in computer vision, including object localization~\cite{yu2022object}, object detection~\cite{chen2021points,wang2022omni}, image instance segmentation ~\cite{laradji2020proposal,cheng2021pointly}, and image panoptic segmentation~\cite{li2022fully,fan2022pointly}. Point-level interactions are also popular in the field of interactive image segmentation~\cite{chen2022focalclick,liu2022pseudoclick,cheng2021modular} and interactive video segmentation~\cite{price2009livecut,benard2017interactive,caelles20182018,heo2021guided}, with a focus on label propagation or reducing interaction time. The closest work to ours is PointPanoptic~\cite{fan2022pointly} where they use a single point to train image panoptic segmentation. Note that this fundamentally differs from our work, as we aim for reducing video annotations by utilizing pretrained image representations while PointPanoptic~\cite{fan2022pointly} aims for reducing image annotations by training from scratch. To our best knowledge, there is no prior work that utilizes point-level supervision for VIS.
\vspace{-1mm}

% Recently point-level supervision has attracted growing attention in computer vision, including object localization~\cite{yu2022object}, object detection~\cite{chen2021points,wang2022omni}, image instance segmentation ~\cite{laradji2020proposal,cheng2021pointly}, and image panoptic segmentation~\cite{li2022fully,fan2022pointly}. Point-level interactions are also popular in the field of interactive image segmentation~\cite{chen2022focalclick,liu2022pseudoclick,cheng2021modular} and interactive video segmentation~\cite{price2009livecut,caelles20182018,heo2021guided}. The closest work to ours is PointPanoptic~\cite{fan2022pointly} where they use a single point to train image panoptic segmentation. Note that this fundamentally differs from our work, as we aim for reducing video annotations by utilizing pretrained image representations while PointPanoptic~\cite{fan2022pointly} aims for reducing image annotations by training from scratch. To our best knowledge, there is no prior work that utilizes point-level supervision for VIS.

\section{Method}
\label{sec:method}

Our \ours design is motivated by three major challenges in point-supervised VIS:

\noindent\textbf{Sparsity in ground truth.} Unlike masks and boxes, points do not provide detailed localization and shape information about objects, especially their boundaries, sizes and extreme points. Points are also sparse, which causes difficulties during model training. Our task is further complicated by the lack of negative supervision when only positive points are annotated, which is important to learn a correct decision boundary. Since many VIS methods involve COCO pretraining, we similarly leverage this pipeline to enrich our supervision using object shape priors.

\noindent\textbf{Matching sparse annotations.} An important step in recent Transformer-based instance segmentation models is to match their mask proposals with ground truth masks using some costs, usually defined as the intersection-over-union (IoU) between two masks. This matching step is the key to enable end-to-end training without non-maximum suppression (NMS). However, such matching process becomes problematic in our case, since points do not provide informative measurement on how accurate a mask is. Therefore, special designs need to be taken into consideration when defining the cost of matching step in \ours.

\noindent\textbf{Novel categories.} In real world applications, there is no guarantee that the categories of downstream VIS tasks overlap with those in images. Therefore, a point-supervised framework has to handle arbitrary new categories and learn efficiently with sparse supervision.

\noindent\textbf{Our solution.} We therefore propose several designs in \ours to address the above challenges. We first pre-train instance segmentation models on COCO and use the pretrained models to generate spatiao-temporal mask proposals in training videoes. Despite the issue mentioned above, this class-agnostic proposal generation mechanism works well in open set scenarios, providing good coverage on categories not seen in COCO. We then propose a point-based matcher that incorporates annotation-free negative cues from other instances in the same video frame. Finally, we address the generalization issue for new categories in videos by conducting self-training to mitigate the domain gap and refine our results. We iterate the training with pseudo masks from prior round. These solutions together allow us to learn VIS with points effectively.

\subsection{Problem Setup}
\label{subsec:setup}

Due to the high annotation cost in videos, in the standard supervised setup, a prevalent strategy to train a video instance segmentation model is to first train on image instance segmentation datasets, and then finetune on video instance segmentation datasets. Formally, there is a full-labeled image instance segmentation dataset $\mathcal{D}_{I}$ with category space size ${\mathbf{O}_I}$ and a full-labeled video instance segmentation dataset $\mathcal{D}_{V}$ with category space size ${\mathbf{O}_V}$, both of which are annotated with object masks.

In our proposed point-supervised setup, we adopt a video dataset with point annotations (denoted as $\mathcal{D}_{V_p}$) instead of with masks for each video object as in $\mathcal{D}_{V}$. Specifically, given a video $\mathbf{V}\in\mathbb{R}^{H\times W\times T}$ of $T$ RGB frames with width $W$ and height $H$, the point annotations for the $j^{\text{th}}$ video object in the video $\mathbf{V}$ is denoted as $\mathbf{G}_j=\{\{\mathbf{P}_j^t,\mathbf{L}_j^t\}_{t=1}^T,\mathbf{b}_j\}$, where $\mathbf{P}_j^t\in\mathbb{R}^{N_j^t\times 2}$ are the x-y coordinates of the annotated points at the $t^{\text{th}}$ frame, $\mathbf{L}_j^t\in\mathbb{R}^{N_j^t}$ are the corresponding binary labels for the annotated points indicating foreground or background, $N_j^t$ is the number of annotated points, $\mathbf{b}_j\in\mathbb{R}^{\mathbf{O}_V}$ is the one-hot category label and $t$ is the time index. In this setup, a video instance segmentation model $\mathbf{F}$ is first trained on image dataset $\mathcal{D}_{I}$ with full masks, resulting an image model $\mathbf{F}(;\theta_I)$. Then $\mathbf{F}(;\theta_I)$ is finetuned on video dataset $\mathcal{D}_{V_p}$ with point-level annotations, resulting the final video instance segmentation model $\mathbf{F}(;\theta_{V_p})$.

Supervising a model solely based on annotated points with a loss function like cross-entropy can lead to the model collapsing, especially if only positive points are annotated. Therefore, densifying point annotations and including negative supervision signals become crucial for learning a structured dense prediction task via points.

% A naive solution to train a model with point annotations is to solely supervise the locations with annotated points, and apply a loss (e.g. cross-entropy) in the same way as full supervision to propagate the gradients. However, learning a structured dense prediction task via points is not easy, especially when only positive points annotated, in which case the model could collapse (as will be shown in our ablation). How to densify the point annotations and include more negative supervision signals becomes a critical question.

\begin{figure*}[th]
  \centering
  \includegraphics[width=1.0\linewidth]{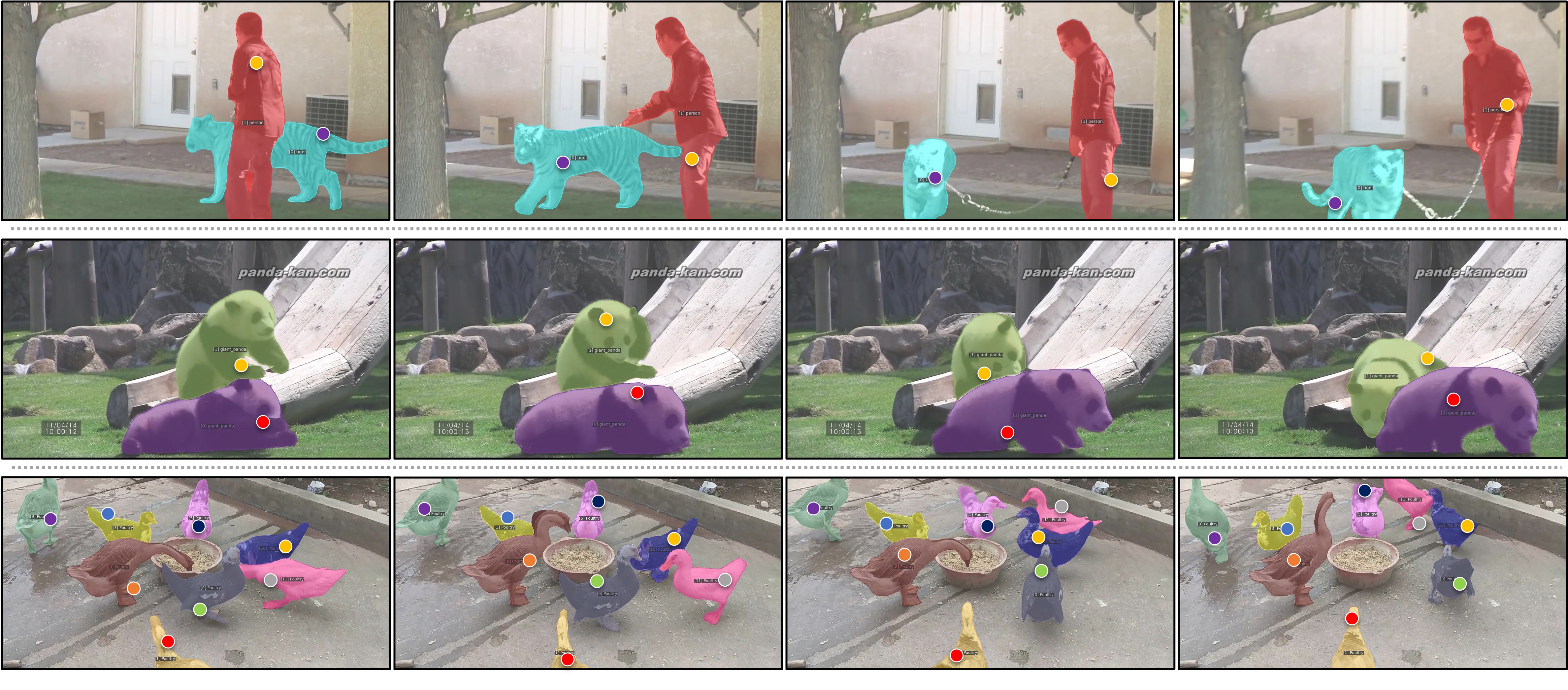}
  \caption{Visualization of point annotations and pseudo masks obtained by our method on Youtube-VIS 2019~\cite{yang2019video} (row1), Youtube-VIS 2021~\cite{yang2019video} (row2), and OVIS~\cite{qi2021occluded} (row3) training set.}
  \label{fig:vis_ann}
\end{figure*}
\subsection{Learning VIS with Sparse Points}
\label{subsec:dense}

% solve two challenges: dense and sparsity
\noindent\textbf{Class-agnostic proposal generation.}
To obtain meaningful dense supervision with abundant negative signals, we shift one step back to the image model instead of dwelling on a degenerated video model finetuned with points. Recall that we have image instance segmentation datasets with mask annotations ready at hands. A pretrained image model on such datasets should already know rough object shape, even if never trained on videos and the categories do not fully overlap. Given this simple yet non-trivial discovery, we propose to generate dense class-agnostic spatial-temporal proposals for each video by utilizing a pretrained image model that encodes rich shape prior.
% which will be used as psuedo-labels? Should explain what the proposals are used for? Too wordy now?

It is challenging to obtain spatio-temporal proposals when video supervision is unavailable. %We address this challenge by adopting an image-based approach with a transformer-based model during training and then links per image proposal by query-matching during inference. 
% This is different from MinVIS as MinVIS is supervised while we use it for a different purpose of generating spatio-temporal pseudo masks in the point-supervised new context.
We address this challenge by leveraging COCO pre-trained image instance segmentation model~\cite{cheng2021masked} to generate per-frame instance proposals and use bipartite matching on query embeddings to convert them to spatio-temporal video instance proposals. While previous work has used a similar technique for VIS with full video supervision~\cite{huang2022minvis}, we here show that this enables cross-domain video instance proposals generation trained with image dataset only.

%While previous work has used a similar technique for fully supervised video instance segmentation~\cite{huang2022minvis}, we show that this enables video instance proposals \emph{without any video supervision}.

% hsy v0.1: todo: not to mention video-based method
% To obtain spatio-temporal proposals for videos from a model trained only on image dataset, we have two potential approaches. The video-based method trains on synthetically generated video sequences given transformed images and can directly produce spatio-temporal proposals during inference. The image-based method trains a transformer-based model on images and then links per-frame proposals by query-matching during inference~\cite{huang2022minvis}. We follow the latter one and refer readers to MinVIS~\cite{huang2022minvis} for more details.

% Concretely, given model $\mathbf{F}(;\theta_I)$ trained on image dataset $\mathcal{D}_I$ and a video sequence $\mathbf{V}$ from $\mathcal{D}_{V}$, we obtain the initial spatio-temporal proposals for $\mathbf{V}$ as follows:
% \begin{align}
% \label{eq:proposal}
%    \{\hat{\mathbf{M}}_r\}_{r=1}^R=\mathbf{F}(\mathbf{V};\theta_I)
% \end{align}, where $\hat{\mathbf{M}}\in\mathbb{R}^{H\times W\times T}$ is a spatial-temporal proposal with continous logits after sigmoid but before binarization, $R$ is the maximum number of proposals for a video (e.g. 100). 

Concretely, given model $\mathbf{F}(;\theta_I)$ trained on image dataset $\mathcal{D}_I$ and a video sequence $\mathbf{V}$ from $\mathcal{D}_{V_p}$, we obtain the initial proposals $\hat{\mathbf{R}}$ for $\mathbf{V}$ by conducting inference as below:
\begin{align}
\label{eq:proposal}
    \hat{\mathbf{R}}=\mathbf{F}(\mathbf{V};\theta_I)=\{\hat{\mathbf{M}}_r,\hat{c}_r\}_{r=1}^R
\end{align}, where $\hat{\mathbf{M}}\in\mathbb{R}^{H\times W\times T}$ is a spatial-temporal proposal with continous logits after sigmoid but before binarization, $\hat{c}_r$ is the confidence score, $R$ is the maximum number of proposals for a video (e.g. 100). 

Given the above dense proposals, there is still one open-set problem worth resolving. As there could be new categories in $\mathcal{D}_{V_p}$ and $\mathbf{F}(;\theta_I)$ has never been finetuned on $\mathcal{D}_{V_p}$, the confidence score $\hat{c}_r$ is not meaningful for every video. To represent the confidence of class-agnostic proposals without the reliance on categories, we propose to use maskness score following~\cite{wang2022freesolo} to obtain the confidence of a mask as $ c_r=\frac{1}{H\times W\times T}\sum_{x,y,z}\hat{\mathbf{M}}_r(x,y,z)$ where $x$,$y$ are the x-axis and y-axis spatial coordinates, $t$ indexes the time.

Therefore, our final class-agnostic dense spatio-temporal proposals $\mathbf{R}$ for a video $\mathbf{V}$ is denoted as $\mathbf{R}=\{\mathbf{M}_r,c_r\}_{r=1}^R$, where $\mathbf{M}_r\in\mathbb{R}^{H\times W\times T}$ is the binarized mask of $\hat{\mathbf{M}}_r$, $c_r$ is the maskness score.

\paragraph{Point-based matcher.}
Given the dense class-agnostic proposals, the next challenge is to assign the best proposal to a video object and produce the final pseudo mask when only point annotations are available in videos. This can be challenging since there may be multiple proposals overlapping with a single point, making it difficult to determine which proposal provides the best boundary information.

To address this issue, we develop a matching cost function that combines cues from both point annotations and spatio-temporal proposals to effectively match proposals and video objects with points. With our proposed point-based matcher, we can formulate the pseudo-label filtering problem as a bipartite matching problem between the proposals and the objects. We provide details on the matching process and the design of the matching cost below.

Specifically, we search for a permutation $\hat{\sigma}$ between the set of dense proposals and the set of video ground truth with points given a video. Assuming $R$ is larger than the number of objects in the video, we consider $\mathbf{G}$ as a set of video ground truth with size $R$ padded with $\varnothing$ (no object). To find a bipartite matching between $\mathbf{R}$ and $\mathbf{G}$, we search for a permutation $\sigma\in \Omega_R$ of $R$ elements with the lowest cost:
\begin{align}
    \hat{\sigma} = \argmin_{\sigma\in\Omega_R} \sum_{j=1}^{R}\mathcal{L}_{\text{match}}(\mathbf{G}_j,\mathbf{R}_{\sigma(j)})
\end{align} where $\mathcal{L}_{\text{match}}(\mathbf{G}_j,\mathbf{R}_{\sigma(j)})$ is a pair-wise matching cost between ground truth $\mathbf{G}_j$ and a proposal with index $\sigma(j)$. This optimal assignment is computed efficiently with the Hungarian algorithm following prior work~\cite{carion2020end}. Next, we present the specific matching cost $\mathcal{L}_{\text{match}}$ given point annotations.

To penalize proposals that do not overlap with the annotated points consistently, we first define an annotated cost $\mathcal{L}_{\text{ann}}(\mathbf{G}_j,\mathbf{R}_{\sigma(j)})$ calculated over $T$ frames as below:
\begin{align}
    \mathcal{L}_{\text{ann}}(\mathbf{G}_j,\mathbf{R}_{\sigma(j)}) = \sum_{t=1}^{T}\sum_{k=1}^{N_j^t}\mathbbm{1}[\mathbf{M}_{\sigma(j)}(\mathbf{P}_j^t(k),t)\neq \mathbf{L}_j^t(k)]
\end{align} where $\mathbbm{1}[\cdot]$ is the indicator function, $k$ is the point index, $t$ is the time index, $\mathbf{P}_j^t(k)\in\mathbb{R}^2$ is the x-y coordinates for the $k^{\text{th}}$ point of the $j^{\text{th}}$ video object at the $t^{\text{th}}$ frame.

To combat a server lack of negative points during matching, especially when only positive points are annotated, we further develop a cross instance negative cost $\mathcal{L}_{\text{cineg}}(\mathbf{G}_j,\mathbf{R}_{\sigma(j)})$ to filter out multiple overlapping proposals. The key idea is that the positive points for one video object can serve as accurate negative points for the other video objects in the same video frame. By aggregating the positive point annotations from other video objects in the same frame, we obtain additional accurate negative point annotations for each video object. Therefore, this annotation-free $\mathcal{L}_{\text{cineg}}$ can penalize inaccurate proposals that overlap with the positively annotated points in other video instances.

In addition to the annotated cost and cross instance negative cost, we also define a maskness cost $\mathcal{L}_{\text{maskness}}$ as the negative of the maskness score to favor confident proposals. As a result, our proposed matching cost $\mathcal{L}_{\text{match}}$ is a weighted combination of the annotated cost, cross instance negative cost, and maskness cost as below:
\begin{align}
\label{eqn:match}
    \mathcal{L}_{\text{match}} = \lambda_1\mathcal{L}_{\text{ann}} +  \lambda_2\mathcal{L}_{\text{cineg}} + \lambda_3\mathcal{L}_{\text{maskness}}
\end{align} where $\lambda_1$, $\lambda_2$ and $\lambda_3$ are the weight balancing parameters. 

To handle the birth and death of objects, we compute the matching cost only over video frames where objects show up. After matching,
we remove the pseudo-labels for frames where objects die. With our carefully designed matching cost, we can obtain high quality dense pseudo-mask for objects with point annotations via the optimal permutation.

%%%%%%%%%%%%%%%%%%%%%%%%%%%Three benchmarks in one table
\begin{table*}[t]
\centering
\resizebox{\textwidth}{!}{
\addtolength{\tabcolsep}{4pt}
\tabfontsize
\begin{tabular}{llllllllll}
\toprule
Method       &Dataset   &Sup. & AP   & AP$_{50}$ & AP$_{75}$ & AR$_1$ & AR$_{10}$ \\\midrule
TeViT~\cite{yang2022tevit}           & YouTube-VIS-2019  &$\mathcal{M}$ & 56.8 & 80.6    & 63.1    & 52.0   & 63.3    \\
IDOL~\cite{IDOL}            & YouTube-VIS-2019 &$\mathcal{M}$& 64.3 & 87.5    & 71.0    & 55.6   & 69.1    \\
MinVIS~\cite{huang2022minvis} &YouTube-VIS-2019 &$\mathcal{M}$ & 61.6 & 83.3    & 68.6    & 54.8   & 66.6    \\
\ours (P1)     &YouTube-VIS-2019    & $\mathcal{P}_1$   & 53.9 (87.5\%) & 75.7 (90.9\%)    & 61.8 (90.1\%)    & 47.5 (86.7\%)   & 61.4 (92.2\%)    \\
\ours (P1N1)     &YouTube-VIS-2019    & $\mathcal{P}_2$   & 59.6 (96.7\%) & 83.3 (100\%)    & 67.1 (97.8\%)    & 52.7 (96.2\%)   & 63.8 (95.8\%)    \\

\midrule
SeqFormer~\cite{wu2021seqformer}       & YouTube-VIS-2021   & $\mathcal{M}$ & 51.8 & 74.6    & 58.2    & 42.8   & 58.1    \\
IDOL~\cite{IDOL}            & YouTube-VIS-2021    & $\mathcal{M}$  & 56.1 & 80.8    & 63.5    & 45.0   & 60.1    \\
MinVIS~\cite{huang2022minvis}   &YouTube-VIS-2021        &$\mathcal{M}$& 55.3 & 76.6    & 62.0    & 45.9   & 60.8    \\
\ours(P1)     &YouTube-VIS-2021       & $\mathcal{P}_1$  & 46.3 (83.7\%) & 70.5 (92.0\%)    & 51.1 (82.4\%)    & 37.7 (82.1\%)   & 52.9 (87.0\%)    \\
\ours(P1N1)    &YouTube-VIS-2021     &$\mathcal{P}_2$  & 48.5 (87.7\%) & 73.0 (95.3\%)    & 54.4 (87.7\%)   & 41.7 (90.8\%)   & 54.1 (89.0\%)    \\

\midrule
 
MaskTrack~\cite{li2021limited} & Occluded VIS         &$\mathcal{M}$  &28.9 &56.3& 26.8& 13.5& 34.0 \\
IDOL~\cite{IDOL} & Occluded VIS         &$\mathcal{M}$  &42.6&	65.7&	45.2&			17.9&		49.6 \\
MinVIS~\cite{huang2022minvis}         & Occluded VIS         &$\mathcal{M}$   & 39.4 & 61.5  & 41.3      & 18.1   & 43.3      \\
\ours (P1)                 &  Occluded VIS   &$\mathcal{P}_1$ & 28.6 (72.6\%) & 49.6 (80.7\%) & 27.5 (66.6\%)      & 15.0 (82.9\%)   & 32.8 (75.8\%)      \\
\ours (P1N1)            & Occluded VIS  &$\mathcal{P}_2$ & 28.6 (72.6\%) & 51.2 (83.3\%) & 27.2 (65.9\%)     & 14.7 (81.2\%)  & 32.2 (74.4\%)      \\
\bottomrule
\end{tabular}
}

\caption{\textbf{Full mask ($\mathcal{M}$) vs. our point supervision ($\mathcal{P}$) on validation set of YouTube-VIS 2019~\cite{yang2019video}, YouTube-VIS 2021~\cite{yang2019video}, and OVIS~\cite{qi2021occluded}.} All results below are based on Swin-L backbone. Our PointVIS results are with self-training.}

\label{tab:main}
\end{table*}
%%%%%%%%%%%%%%%%%%%%%%%%%%%%%%%%%%%%%%%%%%%%%%%%%%%%%%%%%%

\paragraph{Self-training for generalization.}
With the above high-quality pseudo masks, we can train our video instance segmentation model on videos with standard loss for mask prediction (cross-entropy and dice loss) and cross-entropy loss for classification following the existing work~\cite{huang2022minvis,cheng2021mask2former}.

To generalize from images to videos for new categories, we conduct self-training by regenerating pseudo-labels again from our finetuned video model. The reason is that our pseudo-labels are initially generated from an image model that has never been trained on videos, and there is obviously a domain gap. During self-training, we use confidence score instead of maskness score for pseudo-label matching as the model has been finetuned on videos.

\section{Experiments}
\label{sec:exp}

We evaluate our method on three video instance segmentation benchmarks: YouTube-VIS 2019~\cite{yang2019video}, YouTube-VIS 2021~\cite{yang2019video}, and Occluded VIS~\cite{qi2021occluded}. We describe our experimental setup in Sec.\ref{sec_exp:setup}, compare \ours with state-of-the-art fully-supervised approaches in Sec.\ref{sec_exp:compare}, and provide an ablation study in Sec.~\ref{sec_exp:ablation}. For more details, please refer to the supplementary material.

% We evaluate our method on the video instance segmentation task on three public benchmarks: YouTube-VIS 2019~\cite{yang2019video}, YouTube-VIS 2021~\cite{yang2019video} and Occluded VIS (OVIS)~\cite{qi2021occluded}. In the following sections, we first elaborate on the experimental setup of our proposed method in Sec.~\ref{sec_exp:setup}. Then, we provide the quantitative comparison of \ours with  state-of-the-art fully-supervised approaches in Sec.~\ref{sec_exp:compare}, and follow that with detailed ablation study in Sec.~\ref{sec_exp:ablation}. For more detailed results and analysis, we refer the readers to the supplementary material.

\subsection{Experimental Setup}
\label{sec_exp:setup}

\noindent\textbf{Datasets.} \textbf{YouTube-VIS 2019}~\cite{yang2019video} is a popular dataset for VIS with 2,883 labeled videos, 131K instance masks, and 40 classes. \textbf{YouTube-VIS 2021}~\cite{yang2019video} is an improved version with 8,171 unique video instances and 232k instance masks. \textbf{OVIS}~\cite{qi2021occluded} is a recently proposed challenging VIS dataset with heavy occlusion and long sequences, containing 296k instance masks and 5.8 instance per video from 25 classes. We synthesize point annotations by randomly sampling points given the ground truth mask in each frame.

% \textbf{YouTube-VIS 2019}~\cite{yang2019video} is the most commonly used dataset for video instance segmentation, which contains 2,883 videos labeled at 6 FPS, 131K instance masks, and 40 classes. \textbf{YouTube-VIS 2021}~\cite{yang2019video}  is an improved and augmented version of YouTube-VIS-2019 dataset, which has 8,171 unique video instances and 232k instance masks. There are 2985/421/453 videos for train/val/test. \textbf{OVIS}~\cite{qi2021occluded} dataset is a recently proposed challenging VIS dataset with heavy occlusion and long sequences. OVIS consists of 296k instance masks and 5.80 instance per video from 25 classes. There are 607/140/154 videos for train/val/test. We synthesize the point
% annotations for each object by randomly sampling points given the ground truth mask in each frame.

\vspace{1mm}
\noindent\textbf{Architecture and optimization.} We build our \ours on top of MinVIS~\cite{huang2022minvis} by strictly following its model architecture, training hyper parameters and losses. The only two modifications during video training are: 1) we use pseudo masks obtained from our method given point-level annotations while MinVIS uses annotated masks; 2) we use larger iterations as pseudo-labels require longer time for convergence. All models are pre-trained with COCO instance segmentation~\cite{lin2014microsoft} then finetuned on videos with Swin backbone~\cite{liu2021swin} unless otherwise stated. For our point-based matcher, we set $\lambda_1 = 5.0$, $\lambda_2=5.0$, and $\lambda_3=2.0$ for all three datasets. We conduct one iteration of self-training.

%We do not use self-training described in section~\ref{subsec:dense} unless otherwise stated.

\vspace{1mm}
\noindent\textbf{Baselines.} 
We propose new baselines for comparison as no prior work is directly applicable to our new point-supervised setting for VIS. 1) \textbf{VISP}: naive training MinVIS with points, where only locations with annotated points are supervised during video training. 2) \textbf{VISP+CINeg}: adding annotation-free negative point loss to VISP by enforcing cross-instance negative cues on top of it, as described in Sec.~\ref{subsec:dense}. This strategy is denoted as ``CINeg''. 3) \textbf{VISC}: to ablate the benefits of points and decouple the impact of pretrained image model, we obtain pseudo-labels by selecting the top k most confident proposals from our proposal set $\hat{\mathbf{R}}$ sorted by the confidence score $\hat{c}_r$.

% To our best knowledge, there is no prior work directly applicable to our new point-supervised setting for VIS. Therefore we propose several new baselines for direct comparison as below. 1) \textbf{VISP}: to understand what happens when naively training with points, we only supervise MinVIS in locations with annotated points during video training. 2) \textbf{VISP+CINeg}: to understand the benefits of our proposed cross instance negative cues, we enforce additional negative point loss on top of VISP where the negative supervision is obtained in the same way as described in Sec.~\ref{subsec:dense}. We denote this strategy of adding annotation-free negative point loss as cross-instance negatives ``CINeg''. 3) \textbf{VISC}: to ablate the benefits of points and decouple the impact of pretrained image model, we obtain pseudo-labels by directly selecting the top k most confident proposals from our proposal set $\hat{\mathbf{R}}$ sorted by the confidence score $\hat{c}_r$ (c.f. Eq.~\ref{eq:proposal}).  

We also include the existing unsupervised/semi-/weakly-supervised VIS work to compare with our point-supervised setting as follows. 1) \textbf{VIS-Unsup}~\cite{fu2021learning}: pretrained COCO image model built on top of SOLO~\cite{wang2020solo} without finetuning on videos. 2) \textbf{VIS-Semi}~\cite{fu2021learning}: finetuning on videos on top of VIS-Unsup but without video annotations. Note that VIS-Unsup and VIS-Semi can only handle the overlapped categories, therefore we report AP/AR (seen) in line with~\cite{fu2021learning}. 3) \textbf{VIS-Weak}~\cite{liu2021weakly}: the first weakly-supervised VIS model that uses category annotations only in each video frame.

\vspace{1mm}
\noindent\textbf{Metrics.} We use AP and AR for evaluation and train on the training split and evaluate on the validation split using public evaluation servers of the three datasets. Baselines VISC, VIS-Unsup~\cite{fu2021learning} , and VIS-Semi~\cite{fu2021learning} can only handle categories overlapped with the image model, so we report AP/AR (seen) computed by averaging only over the overlapped categories for these baselines following~\cite{fu2021learning}.

% We use Average Precision (AP) and Average Recall (AR) for evaluation~\cite{yang2019video}. We use training split for training and evaluate on the validation split by submitting inference results to the public evaluation servers of the three datasets. Note that the baselines VISC, VIS-Unsup~\cite{fu2021learning} and VIS-Semi~\cite{fu2021learning} could only handle the categories overlapped with the image model due to using confidence score or lacking category annotations in videos, therefore we report AP/AR (seen) computed by averaging over only the overlapped categories for these baselines following~\cite{fu2021learning}.

%%%%%%%%%%%%%%%%%%%%%%%%%%%%%%%%%%%%%%%%%%%%Effects of each components
\begin{table}[t]
\footnotesize
\renewcommand{\arraystretch}{1.2}
\renewcommand{\tabcolsep}{6pt}
%"DPPointMatcher" denotes our Dense Pseudo via Point Matcher. "CINeg" denotes cross instance negative.
\centering
\resizebox{1.0\linewidth}{!}{
\begin{tabular}{@{}lccccc@{}}
\toprule                
Model &$\mathcal{L}_{\text{Ann}}$&$\mathcal{L}_{\text{CINeg}}$&$\mathcal{L}_{\text{maskness}}$&Self-Train&mAP\\
\midrule
MinVIS~\cite{huang2022minvis} (Upperbound) & -  & -& -& -& 55.3       \\
\midrule
VISP (P1) &  - & -& -  & -&0.4       \\
VISP+CINeg (P1)&  - & -& -  &-& 9.7 \\
PointVIS (P1)& \cmark\quad & - & - &-&40.4  \\
PointVIS (P1)& \cmark\quad & \cmark\quad & - &-&45.7  \\
PointVIS (P1)& \cmark\quad& \cmark\quad  &\cmark\quad& -&46.0  \\
PointVIS (P1)& \cmark\quad& \cmark\quad & \cmark\quad  &\cmark&47.3\\
\bottomrule
\end{tabular}
}
\caption{\textbf{Effects of each component on YouTube-VIS 2019~\cite{yang2019video} val-dev.}}
\label{tab:ablation_ind}
\end{table}

\begin{table}[t]
\footnotesize
\renewcommand{\arraystretch}{1.2}
\renewcommand{\tabcolsep}{6pt}
\centering
	\resizebox{1.0\linewidth}{!}{
		\begin{tabular}{@{}llll@{}}
			\toprule                
			Model ID & Sampling method for Pos&Sampling method for Neg&mAP \\
			\midrule
			PointVIS (P1) &Random&-& 46.0   \\
			PointVIS (P1) &Distance Transform&-  & 47.1  \\
			\midrule
			PointVIS (P1N1) &Random&Random (In-box)& 48.6   \\
			PointVIS (P1N1) &Random&Random (Out-box-but-in-200$\%$-box)  & 48.0  \\
			\bottomrule
		\end{tabular}
	}
\caption{\textbf{Analysis of point selection bias on YouTube-VIS 2019~\cite{yang2019video} val-dev.}}
\label{tab:ablation_sample}
\end{table}

\begin{table}[t]
  \small
  \centering

    \centering
  	\resizebox{1.0\linewidth}{!}{
		\begin{tabular}{@{}lllllll@{}}
			\toprule                
			Model ID& CINeg&DPPointMatcher&P1 & P10&P1N1 & P5N5             \\
			\midrule
			VISP &-  &-& 0.4 & 0.5& 27.6 & 33.0  \\
			VISP+CINeg &\cmark &-  & 9.7 & 10.0& 45.6 & 48.9  \\
			\midrule
			\ours (Ours) &\cmark &\cmark& 46.0 & 45.9& 48.6 & 49.4   \\
			\bottomrule
		\end{tabular}
	}
	%Negative points are sampled inside ground truth bounding box.
  \caption{\textbf{Effects of additional points on YouTube-VIS 2019~\cite{yang2019video} val-dev.} ``DPPointMatcher'' means Dense Pseudo via Point Matcher. ``CINeg'' means enforcing additional negative point loss.}
  \label{tab:ablation_morepts}
  \vspace{-2mm}
\end{table}

\subsection{Comparison with Fully-Supervised SOTA}
\label{sec_exp:compare}
We compare our \ours with recent fully-supervised methods including IDOL~\cite{IDOL}, MinVIS~\cite{huang2022minvis}, TeViT~\cite{yang2022tevit}, SeqFormer~\cite{wu2021seqformer}, and MaskTrack~\cite{li2021limited}, as shown in~\tabref{main}. Note that MinVIS~\cite{huang2022minvis} serves as the fully-supervised counterpart of  \ours, therefore we compute the retention rate as the performance of \ours divided by the performance of MinVIS. We use $\mathcal{M}$ to indicate full supervision from dense masks, and $\mathcal{P}_n$ to indicate sparse supervision from $n$ points. P1 means only one positive point is labeled. P1N1 means one positive and one negative opint are labeled, and similarly for larger number of points.

\vspace{1mm}
\noindent\textbf{YouTube-VIS 2019.} With one single point per object (P1), \ours achieves 53.9 mAP. With two points per object (P1N1), \ours achieves 59.6 mAP, which is $96.7\%$ of the supervised counterpart and even surpassed recent fully-supervised method TeViT~\cite{yang2022tevit} by nearly 3 AP. These competitive results demonstrate the potential of our point-supervised video instance segmentation framework.

\vspace{1mm}
\noindent\textbf{YouTube-VIS 2021.} On this more challenging dataset, we can still achieve $87.7\%$ performance of the supervised counterpart, reaching 48.5 AP, which is only 3 AP away from recent fully-supervised approach SeqFormer.

\vspace{1mm}
\noindent\textbf{OVIS.} \ours achieves 28.6 AP on this challenging dataset, 72.6\% of the supervised counterpart, and matches previous fully-supervised methods~\cite{li2021limited}. The retention rate is relatively lower compared to the other two datasets, likely due to the quality of spatio-temporal proposals, which degrades due to extremely long sequences in OVIS.

\begin{table}[t]
\centering
\subfloat[Evaluation on seen categories]{
	\resizebox{1.0\linewidth}{!}{
		\begin{tabular}{@{}llllllll@{}}
			\toprule                
			Model&Backbone&Sup. & mAP&$\text{AP}_{50}$ & $\text{AP}_{75}$&$\text{AR}_1$& $\text{AR}_{10}$             \\
			\midrule

			VIS-Unsup~\cite{fu2021learning}   &R50& $\mathcal{V}$ & 23.9& 43.3 & 21.5 &26.7&37.3 \\
			VIS-Semi~\cite{fu2021learning} &R50  & $\mathcal{V}$ & 38.3& 61.1 & 39.8&36.9&44.5  \\
			%\bottomrule
	
			VISC  &R50& $\mathcal{V}$ & 42.0& 62.5 & 47.3&48.7&56.7   \\
			\midrule
			PointVIS (P1-Ours)  &R50& $\mathcal{P}_1$ & 47.0 & 67.4&53.4&44.4&50.9  \\
		
			PointVIS (P1-Ours) &Swin-L&$\mathcal{P}_1$  & 58.6 & 80.3&66.2&54.1&64.2  \\
			\bottomrule
		\end{tabular}
}
}
\qquad\qquad\qquad%
\subfloat[Evaluation on all categories]{
   \resizebox{1.0\linewidth}{!}{
		\begin{tabular}{@{}llllllll@{}}
			\toprule                
			Model&Backbone&Sup. & mAP&$\text{AP}_{50}$ & $\text{AP}_{75}$&$\text{AR}_1$&$\text{AR}_{10}$             \\
			\midrule
            VIS-Weak~\cite{liu2021weakly}         & R50 &$\mathcal{C}$ & 10.5 & 27.2    & 6.2    & 12.3   & 13.6    \\
			\midrule
			\ours (P1)        & R50 &$\mathcal{P}_1$   & 38.5 & 58.9    & 41.2    & 36.4   & 46.4    \\
			PointVIS (P1-Ours) &Swin-L& $\mathcal{P}_1$  & 52.5 & 74.5&59.2&47.2&61.5  \\
			\bottomrule
		\end{tabular}
	}
}
\caption{\textbf{Comparison with baselines on YouTube-VIS 2019~\cite{yang2019video} validation set.}}
\label{tab:ablation_base}
%\vspace{-5mm}
\end{table}

\vspace{-0mm}
\subsection{Ablation Study}
\label{sec_exp:ablation}
In this section, we conduct ablation studies to 1) verify the effectiveness of each individual component of our proposed framework; 2) analyze training with subsampled frames; 3) analyze the effect of number of points; 4) analyze point selection bias and 5) compare with baselines on YouTube-VIS 2019~\cite{yang2019video}. We split the training set of YouTube-VIS 2019~\cite{yang2019video} into train-dev for training and val-dev for testing with Swin-B backbone unless otherwise stated. We do not use self-training by default for simplicity unless otherwise stated. Each ablation is conducted under the same experimental setting for a fair comparison. 

%We refer readers to the supplement for more ablations.

\vspace{-3mm}
\subsubsection{Effects of model components.}
\tabref{ablation_ind} presents the ablation results for each component of our model. The VISP baseline, which trains the VIS model with only point supervision, yields a near-zero AP. Incorporating cross instance negatives (VISP+CINeg) improves it to 9.7 mAP. Our method achieves a decent mAP of 40.4 by incorporating pseudo masks generated through our point-based matcher with only $\mathcal{L}_{\text{Ann}}$ in the matching process. Further addition of $\mathcal{L}_{\text{CINeg}}$ and $\mathcal{L}_{\text{maskness}}$ results in a significant boost of 5.7 mAP, highlighting the importance of incorporating cross instance negatives. Finally, with self-training reducing the domain gap, our \ours achieves 47.3 mAP, which is 85.6\% of the fully-supervised upperbound.

\vspace{-3mm}
\subsubsection{Analysis of subsampling frames.}
PointVIS can be extended to the setting of using a subset of frames following MinVIS~\cite{huang2022minvis} (Table~\ref{R1_minVIS_subsample}). With point labels in only 1\% frames, PointVIS (55.0 AP) is only 6.6 AP behind the one uses full-masks in 100\% frames, and achieves a retention rate of 93.2\% of its counterpart with full-masks in $1\%$ frames. This indicates the practical utility of our work. 

PointVIS could be further easily extended to a per-frame setting where frames are annotated with points in parallel w/o tracking annotations. The main change here is that we cannot compute spatio-temporal matching cost. Instead, we change matching to per-frame by computing the matching cost for each annotated frame independently. We refer the resulting per-frame model as PointVIS (PF). PointVIS (PF) achieves competitive results while PointVIS has its own advantages, especially with less frames (Table~\ref{R1_minVIS_subsample}).

\vspace{-3mm}
\subsubsection{Analysis of point selection bias.}
To analyze how point selection bias affects performance, we synthesized point annotations using different sampling methods (\tabref{ablation_sample}). Under the P1 setting, we compared random sampling with human distance transform and found that different methods achieved similar results, but distance transform performed around 1AP better. This suggests that our method is generally robust to annotated point locations, but human annotation can potentially yield higher gains. Under the P1N1 setting, we compared sampling negatives inside versus outside the bounding box and found that the latter was only 0.6 AP behind, indicating that negatives do not need to be constrained within boxes. 
% To understand how point selection bias affects the performance, we synthesize point annotations by using different point sampling methods and summarize the ablation results in \tabref{ablation_sample}. We first study the positive points selection bias by comparing random sampling with human distance transform under P1 setting. We observe that different sampling methods achieve comparable results, while distance transform is around 1AP higher. This result show that our method is generally robust to the annotated point location while the point supervision obtained from human annotators could potential give higher gain. We further study the negative points selection bias under P1N1 setting by comparing sampling negatives inside the bounding box versus outside but within the enlarged bounding box. We find that the latter one is only 0.6 AP behinds, which indicates that negatives do not have to be constraint within boxes. We refer reader for more results in the supplement.

\vspace{-3mm}
\subsubsection{Analysis with additional points.}
\tabref{ablation_morepts} summarizes our results on adding more points. Increasing the number of positive points alone (P1 vs P10) did not improve performance because the model lacks additional cues about the background region. Adding both negative and positive points improved the performance of VISP and VISP+CINeg, which rely heavily on point annotations for knowing foreground and background. Our proposed method's performance improved significantly with just one more negative point and remained stable across all point settings, indicating its robustness.

% \tabref{ablation_morepts} summarizes the results of adding more points. We observe that only increasing the number of positive points could not really improve the performance (P1 vs P10). The reason is that the model has no additional cues about where the background region is. In addition, we observe that adding both negative and positive indeed increases the performance of baselines VISP and VISP+CINeg, as they heavily rely on point annotations for knowing foreground and background. We find that the performance of our proposed method increases clearly with just one more negative point, and is general stable across all point settings, which indicates the robustness of our proposed method.

% Treating the ground truth mask as a set of points, we implemented an unpperbound model of our PointVIS. The upperbound performance (50.0\% mAP) does not match the fully-supervised counterpart (55.3\% mAP), as it is bounded by proposals quality. Our PointVIS instead could approach this upperbound with very little point supervision (saturated at 49.5\% mAP w/ P10N10). 

\vspace{-3mm}
\subsubsection{Comparison with additional baselines.}
Table~\ref{tab:ablation_base} shows the comparison with additional baselines. Compared with VIS-Unsup~\cite{fu2021learning} without using video annotations, our \ours outperforms VIS-Unsup by more than 10 AP with little annotation overhead. Compared with the VISC baseline that does not use video annotations but generating pseudo-labels by confidence, \ours outperforms it by 5 AP showing the benefit of point annotations.

\begin{table}[t]
\footnotesize
    \centering
    \vspace{-0.9em}
    \renewcommand{\tabcolsep}{0.7em}
     \resizebox{\linewidth}{!}{
    \begin{tabular}{@{}l ll c c c c @{}}
    \toprule
     Model &Sup. &Matching& 1\%  & 5\% & 10\% & 100\%\\
        \midrule
        MinVIS [19] &$\mathcal{M}$ & -& 59.0    &59.3& 61.0  & 61.6 \\
        \midrule
        PointVIS (PF) &$\mathcal{P}_2$ &Per-frame& \makecell{52.3 \\($88.6\%$)} & \makecell{54.5 \\($91.9\%$)} & \makecell{54.7 \\($89.7\%$)} & \makecell{57.3 \\($93.0\%$)}\\
        PointVIS & $\mathcal{P}_2$ &Spatio-temporal& \makecell{55.0 \\($93.2\%$)} & \makecell{55.5 \\($93.6\%$)} & \makecell{56.1 \\($92.0\%$)} & \makecell{57.4 \\ ($93.2\%$)}\\
         \bottomrule
    \end{tabular}
    }
      \caption{\textbf{PointVIS (P1N1) with subsampled video frames on YouTube-VIS 2019 validation set (w/o self-training).}}
    \label{R1_minVIS_subsample}
%\vspace{-5mm}
\end{table}

\label{sec:conclusion}
\begin{figure}[t]
  \centering
  \vspace{0.1cm}
  \includegraphics[width=0.95\linewidth]{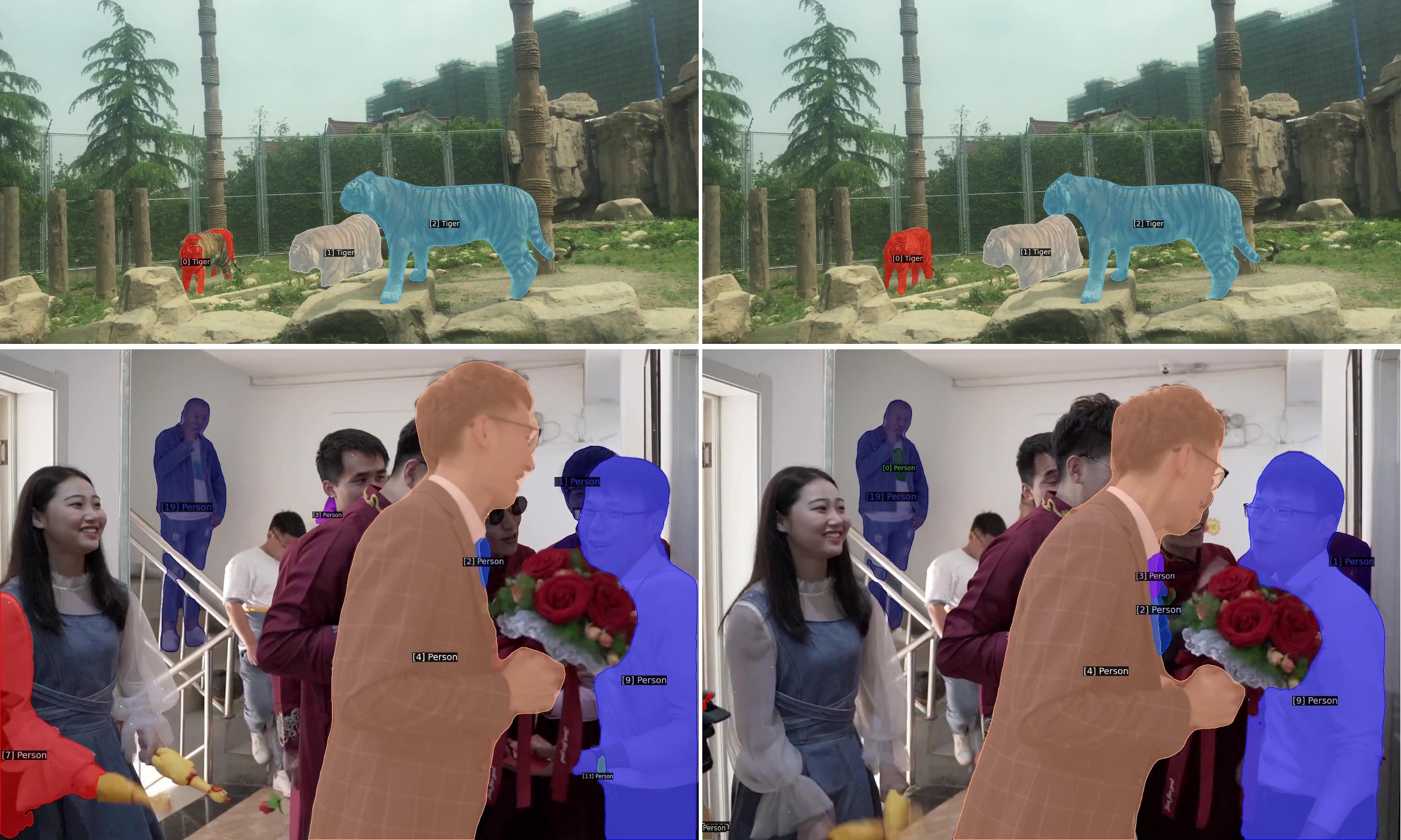}
%   \caption{\textbf{Failure cases.} We visualize our pseudo masks on OVIS~\cite{qi2021occluded} training split. We observe temporal inconsistency (\eg tiger in the top left) or missing instances (\eg person in white).}
 \caption{\textbf{Failure cases on OVIS~\cite{qi2021occluded}.} We observe temporal inconsistency (\eg tiger in the top left) or missing instances (\eg person in white) in our pseudo masks.}
  \label{fig:vis_fail}
 \vspace{-5mm}
\end{figure}

\vspace{-0mm}
\section{Conclusion}
In this work, we address the challenging point-supervised video instance segmentation problem and introduce a point-supervised VIS framework.  The key ingredients are utilizing object shape prior from a pretrianed COCO image instance segmentation model, and our proposed spatio-temporal point-based matcher for generating high-quality dense pseudo-labels for videos. Our method allows us to reduce annotations to
only one point for each object in a video frame, yet retaining
high quality mask predictions close to full supervision. We conduct comprehensive experiments on three datasets and achieve competitive performance compared with the fully-supervised methods.

%\vspace{1mm}
\noindent\textbf{Limitations.} While \ours shows promising results in the direction of weakly-supervised video instance segmentation, it has some limitations. Typical failure cases are shown in \figref{vis_fail}. We observe missing instances and temporal inconsistency. One hypothesis is that our performance is bounded by the quality of proposals, and we believe stronger video instance architecture can mitigate this gap. Another potential direction is to utilize video correspondences for label propagation and other denoising techniques for higher-quality proposals. We leave these as future work.

%\vspace{1mm}
\noindent\textbf{Broader Impacts.} We open up the possibility of learning video instance segmentation with point-level supervision. We hope that our framework can be used to largely reduce annotation cost for a wide range of video recognition tasks in computer vision, including video object detection, video object segmentation, and video panoptic segmentation.

% \noindent\textbf{Broader Impacts.} We open up the possibility of learning video instance segmentation with point-level supervision. We hope that our proposed framework can be used to significantly reduce annotation cost for a wide range of video recognition tasks in computer vision, including video object detection, video object segmentation, and video panoptic segmentation.

{
    \small
    \bibliographystyle{ieeenat_fullname}
    \bibliography{vis}

\begin{thebibliography}{66}
\providecommand{\natexlab}[1]{#1}
\providecommand{\url}[1]{\texttt{#1}}
\expandafter\ifx\csname urlstyle\endcsname\relax
  \providecommand{\doi}[1]{doi: #1}\else
  \providecommand{\doi}{doi: \begingroup \urlstyle{rm}\Url}\fi

\bibitem[Ahn et~al.(2019)Ahn, Cho, and Kwak]{ahn2019weakly}
Jiwoon Ahn, Sunghyun Cho, and Suha Kwak.
\newblock Weakly supervised learning of instance segmentation with inter-pixel
  relations.
\newblock In \emph{CVPR}, 2019.

\bibitem[Arun et~al.(2020)Arun, Jawahar, and Kumar]{arun2020weakly}
Aditya Arun, CV Jawahar, and M~Pawan Kumar.
\newblock Weakly supervised instance segmentation by learning annotation
  consistent instances.
\newblock In \emph{European Conference on Computer Vision}, pages 254--270.
  Springer, 2020.

\bibitem[Athar et~al.(2020)Athar, Mahadevan, Osep, Leal-Taix{\'e}, and
  Leibe]{athar2020stem}
Ali Athar, Sabarinath Mahadevan, Aljosa Osep, Laura Leal-Taix{\'e}, and Bastian
  Leibe.
\newblock Stem-seg: Spatio-temporal embeddings for instance segmentation in
  videos.
\newblock In \emph{ECCV}, 2020.

\bibitem[Bearman et~al.(2016)Bearman, Russakovsky, Ferrari, and
  Fei-Fei]{bearman2016s}
Amy Bearman, Olga Russakovsky, Vittorio Ferrari, and Li Fei-Fei.
\newblock What’s the point: Semantic segmentation with point supervision.
\newblock In \emph{European conference on computer vision}, pages 549--565.
  Springer, 2016.

\bibitem[Benard and Gygli(2017)]{benard2017interactive}
Arnaud Benard and Michael Gygli.
\newblock Interactive video object segmentation in the wild.
\newblock \emph{arXiv preprint arXiv:1801.00269}, 2017.

\bibitem[Bertasius and Torresani(2020)]{bertasius2020classifying}
Gedas Bertasius and Lorenzo Torresani.
\newblock Classifying, segmenting, and tracking object instances in video with
  mask propagation.
\newblock In \emph{CVPR}, 2020.

\bibitem[Bolya et~al.(2019)Bolya, Zhou, Xiao, and Lee]{bolya2019yolact}
Daniel Bolya, Chong Zhou, Fanyi Xiao, and Yong~Jae Lee.
\newblock Yolact: Real-time instance segmentation.
\newblock In \emph{Proceedings of the IEEE/CVF International Conference on
  Computer Vision}, pages 9157--9166, 2019.

\bibitem[Caelles et~al.(2018)Caelles, Montes, Maninis, Chen, Van~Gool, Perazzi,
  and Pont-Tuset]{caelles20182018}
Sergi Caelles, Alberto Montes, Kevis-Kokitsi Maninis, Yuhua Chen, Luc Van~Gool,
  Federico Perazzi, and Jordi Pont-Tuset.
\newblock The 2018 davis challenge on video object segmentation.
\newblock \emph{arXiv preprint arXiv:1803.00557}, 2018.

\bibitem[Cao et~al.(2020)Cao, Anwer, Cholakkal, Khan, Pang, and
  Shao]{cao2020sipmask}
Jiale Cao, Rao~Muhammad Anwer, Hisham Cholakkal, Fahad~Shahbaz Khan, Yanwei
  Pang, and Ling Shao.
\newblock Sipmask: Spatial information preservation for fast image and video
  instance segmentation.
\newblock In \emph{Computer Vision--ECCV 2020: 16th European Conference,
  Glasgow, UK, August 23--28, 2020, Proceedings, Part XIV 16}, pages 1--18.
  Springer, 2020.

\bibitem[Carion et~al.(2020)Carion, Massa, Synnaeve, Usunier, Kirillov, and
  Zagoruyko]{carion2020end}
Nicolas Carion, Francisco Massa, Gabriel Synnaeve, Nicolas Usunier, Alexander
  Kirillov, and Sergey Zagoruyko.
\newblock End-to-end object detection with transformers.
\newblock In \emph{ECCV}, 2020.

\bibitem[Chen et~al.(2021)Chen, Yang, Zhang, Zhang, and Sun]{chen2021points}
Liangyu Chen, Tong Yang, Xiangyu Zhang, Wei Zhang, and Jian Sun.
\newblock Points as queries: Weakly semi-supervised object detection by points.
\newblock In \emph{Proceedings of the IEEE/CVF Conference on Computer Vision
  and Pattern Recognition}, pages 8823--8832, 2021.

\bibitem[Chen et~al.(2022)Chen, Zhao, Zhang, Duan, Qi, and
  Zhao]{chen2022focalclick}
Xi Chen, Zhiyan Zhao, Yilei Zhang, Manni Duan, Donglian Qi, and Hengshuang
  Zhao.
\newblock Focalclick: Towards practical interactive image segmentation.
\newblock In \emph{Proceedings of the IEEE/CVF Conference on Computer Vision
  and Pattern Recognition}, pages 1300--1309, 2022.

\bibitem[Cheng et~al.(2021{\natexlab{a}})Cheng, Choudhuri, Misra, Kirillov,
  Girdhar, and Schwing]{cheng2021mask2former}
Bowen Cheng, Anwesa Choudhuri, Ishan Misra, Alexander Kirillov, Rohit Girdhar,
  and Alexander~G Schwing.
\newblock Mask2former for video instance segmentation.
\newblock \emph{arXiv preprint arXiv:2112.10764}, 2021{\natexlab{a}}.

\bibitem[Cheng et~al.(2021{\natexlab{b}})Cheng, Parkhi, and
  Kirillov]{cheng2021pointly}
Bowen Cheng, Omkar Parkhi, and Alexander Kirillov.
\newblock Pointly-supervised instance segmentation.
\newblock \emph{arXiv}, 2021{\natexlab{b}}.

\bibitem[Cheng et~al.(2022)Cheng, Misra, Schwing, Kirillov, and
  Girdhar]{cheng2021masked}
Bowen Cheng, Ishan Misra, Alexander~G Schwing, Alexander Kirillov, and Rohit
  Girdhar.
\newblock Masked-attention mask transformer for universal image segmentation.
\newblock In \emph{CVPR}, 2022.

\bibitem[Cheng et~al.(2021{\natexlab{c}})Cheng, Tai, and
  Tang]{cheng2021modular}
Ho~Kei Cheng, Yu-Wing Tai, and Chi-Keung Tang.
\newblock Modular interactive video object segmentation: Interaction-to-mask,
  propagation and difference-aware fusion.
\newblock In \emph{Proceedings of the IEEE/CVF Conference on Computer Vision
  and Pattern Recognition}, pages 5559--5568, 2021{\natexlab{c}}.

\bibitem[Fan et~al.(2022)Fan, Zhang, and Tan]{fan2022pointly}
Junsong Fan, Zhaoxiang Zhang, and Tieniu Tan.
\newblock Pointly-supervised panoptic segmentation.
\newblock In \emph{European Conference on Computer Vision}, pages 319--336.
  Springer, 2022.

\bibitem[Fu et~al.(2020)Fu, Yang, Liu, Huang, and Shi]{fu2020compfeat}
Yang Fu, Linjie Yang, Ding Liu, Thomas~S Huang, and Humphrey Shi.
\newblock Compfeat: Comprehensive feature aggregation for video instance
  segmentation.
\newblock \emph{arXiv preprint arXiv:2012.03400}, 2020.

\bibitem[Fu et~al.(2021)Fu, Liu, Iqbal, De~Mello, Shi, and
  Kautz]{fu2021learning}
Yang Fu, Sifei Liu, Umar Iqbal, Shalini De~Mello, Humphrey Shi, and Jan Kautz.
\newblock Learning to track instances without video annotations.
\newblock In \emph{CVPR}, 2021.

\bibitem[Goel et~al.(2020)Goel, Kanazawa, and Malik]{goel2020shape}
Shubham Goel, Angjoo Kanazawa, and Jitendra Malik.
\newblock Shape and viewpoint without keypoints.
\newblock In \emph{European Conference on Computer Vision}, pages 88--104.
  Springer, 2020.

\bibitem[He et~al.(2017)He, Gkioxari, Doll{\'a}r, and Girshick]{he2017mask}
Kaiming He, Georgia Gkioxari, Piotr Doll{\'a}r, and Ross Girshick.
\newblock Mask r-cnn.
\newblock In \emph{ICVV}, 2017.

\bibitem[Heo et~al.(2021)Heo, Koh, and Kim]{heo2021guided}
Yuk Heo, Yeong~Jun Koh, and Chang-Su Kim.
\newblock Guided interactive video object segmentation using reliability-based
  attention maps.
\newblock In \emph{Proceedings of the IEEE/CVF Conference on Computer Vision
  and Pattern Recognition}, pages 7322--7330, 2021.

\bibitem[Hsu et~al.(2019)Hsu, Hsu, Tsai, Lin, and Chuang]{hsu2019weakly}
Cheng-Chun Hsu, Kuang-Jui Hsu, Chung-Chi Tsai, Yen-Yu Lin, and Yung-Yu Chuang.
\newblock Weakly supervised instance segmentation using the bounding box
  tightness prior.
\newblock \emph{Advances in Neural Information Processing Systems},
  32:\penalty0 6586--6597, 2019.

\bibitem[Hu et~al.(2017)Hu, Lan, Jiang, Cao, and Sha]{hu2017fastmask}
Hexiang Hu, Shiyi Lan, Yuning Jiang, Zhimin Cao, and Fei Sha.
\newblock Fastmask: Segment multi-scale object candidates in one shot.
\newblock In \emph{Proceedings of the IEEE Conference on Computer Vision and
  Pattern Recognition}, pages 991--999, 2017.

\bibitem[Huang et~al.(2022)Huang, Yu, and Anandkumar]{huang2022minvis}
De-An Huang, Zhiding Yu, and Anima Anandkumar.
\newblock Minvis: A minimal video instance segmentation framework without
  video-based training.
\newblock \emph{arXiv preprint arXiv:2208.02245}, 2022.

\bibitem[Hwang et~al.(2021)Hwang, Heo, Oh, and Kim]{hwang2021video}
Sukjun Hwang, Miran Heo, Seoung~Wug Oh, and Seon~Joo Kim.
\newblock Video instance segmentation using inter-frame communication
  transformers.
\newblock \emph{NeurIPS}, 2021.

\bibitem[Kanazawa et~al.(2018)Kanazawa, Tulsiani, Efros, and
  Malik]{kanazawa2018learning}
Angjoo Kanazawa, Shubham Tulsiani, Alexei~A Efros, and Jitendra Malik.
\newblock Learning category-specific mesh reconstruction from image
  collections.
\newblock In \emph{Proceedings of the European Conference on Computer Vision
  (ECCV)}, pages 371--386, 2018.

\bibitem[Kanazawa et~al.(2019)Kanazawa, Zhang, Felsen, and
  Malik]{humanMotionKanazawa19}
Angjoo Kanazawa, Jason~Y. Zhang, Panna Felsen, and Jitendra Malik.
\newblock Learning 3d human dynamics from video.
\newblock In \emph{Computer Vision and Pattern Regognition (CVPR)}, 2019.

\bibitem[Lan et~al.(2021)Lan, Yu, Choy, Radhakrishnan, Liu, Zhu, Davis, and
  Anandkumar]{lan2021discobox}
Shiyi Lan, Zhiding Yu, Christopher Choy, Subhashree Radhakrishnan, Guilin Liu,
  Yuke Zhu, Larry~S Davis, and Anima Anandkumar.
\newblock Discobox: Weakly supervised instance segmentation and semantic
  correspondence from box supervision.
\newblock In \emph{ICCV}, 2021.

\bibitem[Laradji et~al.(2020)Laradji, Rostamzadeh, Pinheiro, Vazquez, and
  Schmidt]{laradji2020proposal}
Issam~H Laradji, Negar Rostamzadeh, Pedro~O Pinheiro, David Vazquez, and Mark
  Schmidt.
\newblock Proposal-based instance segmentation with point supervision.
\newblock In \emph{2020 IEEE International Conference on Image Processing
  (ICIP)}, pages 2126--2130. IEEE, 2020.

\bibitem[Lee and Park(2020)]{lee2020centermask}
Youngwan Lee and Jongyoul Park.
\newblock Centermask: Real-time anchor-free instance segmentation.
\newblock In \emph{Proceedings of the IEEE/CVF conference on computer vision
  and pattern recognition}, pages 13906--13915, 2020.

\bibitem[Li et~al.(2021{\natexlab{a}})Li, Li, Li, and Zhang]{li2021spatial}
Minghan Li, Shuai Li, Lida Li, and Lei Zhang.
\newblock Spatial feature calibration and temporal fusion for effective
  one-stage video instance segmentation.
\newblock In \emph{Proceedings of the IEEE/CVF Conference on Computer Vision
  and Pattern Recognition}, pages 11215--11224, 2021{\natexlab{a}}.

\bibitem[Li et~al.(2017)Li, Qi, Dai, Ji, and Wei]{li2017fully}
Yi Li, Haozhi Qi, Jifeng Dai, Xiangyang Ji, and Yichen Wei.
\newblock Fully convolutional instance-aware semantic segmentation.
\newblock In \emph{Proceedings of the IEEE conference on computer vision and
  pattern recognition}, pages 2359--2367, 2017.

\bibitem[Li et~al.(2022)Li, Zhao, Qi, Chen, Qi, Wang, Li, Sun, and
  Jia]{li2022fully}
Yanwei Li, Hengshuang Zhao, Xiaojuan Qi, Yukang Chen, Lu Qi, Liwei Wang, Zeming
  Li, Jian Sun, and Jiaya Jia.
\newblock Fully convolutional networks for panoptic segmentation with
  point-based supervision.
\newblock \emph{IEEE Transactions on Pattern Analysis and Machine
  Intelligence}, 2022.

\bibitem[Li et~al.(2021{\natexlab{b}})Li, Cao, and Wang]{li2021limited}
Zhuang Li, Leilei Cao, and Hongbin Wang.
\newblock Limited sampling reference frame for masktrack r-cnn.
\newblock In \emph{ICCVW}, 2021{\natexlab{b}}.

\bibitem[Lin et~al.(2021)Lin, Wu, Liu, Lu, and Jia]{lin2021video}
Huaijia Lin, Ruizheng Wu, Shu Liu, Jiangbo Lu, and Jiaya Jia.
\newblock Video instance segmentation with a propose-reduce paradigm.
\newblock \emph{arXiv preprint arXiv:2103.13746}, 2021.

\bibitem[Lin et~al.(2014)Lin, Maire, Belongie, Hays, Perona, Ramanan,
  Doll{\'a}r, and Zitnick]{lin2014microsoft}
Tsung-Yi Lin, Michael Maire, Serge Belongie, James Hays, Pietro Perona, Deva
  Ramanan, Piotr Doll{\'a}r, and C~Lawrence Zitnick.
\newblock Microsoft coco: Common objects in context.
\newblock In \emph{ECCV}, 2014.

\bibitem[Liu et~al.(2021{\natexlab{a}})Liu, Cui, Tan, and Chen]{liu2021sg}
Dongfang Liu, Yiming Cui, Wenbo Tan, and Yingjie Chen.
\newblock Sg-net: Spatial granularity network for one-stage video instance
  segmentation.
\newblock In \emph{Proceedings of the IEEE/CVF Conference on Computer Vision
  and Pattern Recognition}, pages 9816--9825, 2021{\natexlab{a}}.

\bibitem[Liu et~al.(2021{\natexlab{b}})Liu, Ramanathan, Mahajan, Yuille, and
  Yang]{liu2021weakly}
Qing Liu, Vignesh Ramanathan, Dhruv Mahajan, Alan Yuille, and Zhenheng Yang.
\newblock Weakly supervised instance segmentation for videos with temporal mask
  consistency.
\newblock In \emph{CVPR}, 2021{\natexlab{b}}.

\bibitem[Liu et~al.(2022)Liu, Zheng, Planche, Karanam, Chen, Niethammer, and
  Wu]{liu2022pseudoclick}
Qin Liu, Meng Zheng, Benjamin Planche, Srikrishna Karanam, Terrence Chen, Marc
  Niethammer, and Ziyan Wu.
\newblock Pseudoclick: Interactive image segmentation with click imitation.
\newblock \emph{arXiv preprint arXiv:2207.05282}, 2022.

\bibitem[Liu et~al.(2021{\natexlab{c}})Liu, Lin, Cao, Hu, Wei, Zhang, Lin, and
  Guo]{liu2021swin}
Ze Liu, Yutong Lin, Yue Cao, Han Hu, Yixuan Wei, Zheng Zhang, Stephen Lin, and
  Baining Guo.
\newblock Swin transformer: Hierarchical vision transformer using shifted
  windows.
\newblock In \emph{ICCV}, 2021{\natexlab{c}}.

\bibitem[Maag et~al.(2021)Maag, Rottmann, Varghese, H{\"u}ger, Schlicht, and
  Gottschalk]{maag2021improving}
Kira Maag, Matthias Rottmann, Serin Varghese, Fabian H{\"u}ger, Peter Schlicht,
  and Hanno Gottschalk.
\newblock Improving video instance segmentation by light-weight temporal
  uncertainty estimates.
\newblock In \emph{2021 International Joint Conference on Neural Networks
  (IJCNN)}, pages 1--8. IEEE, 2021.

\bibitem[Pinheiro et~al.(2015)Pinheiro, Collobert, and
  Doll{\'a}r]{pinheiro2015learning}
Pedro~O Pinheiro, Ronan Collobert, and Piotr Doll{\'a}r.
\newblock Learning to segment object candidates.
\newblock \emph{arXiv preprint arXiv:1506.06204}, 2015.

\bibitem[Pinheiro et~al.(2016)Pinheiro, Lin, Collobert, and
  Doll{\'a}r]{pinheiro2016learning}
Pedro~O Pinheiro, Tsung-Yi Lin, Ronan Collobert, and Piotr Doll{\'a}r.
\newblock Learning to refine object segments.
\newblock In \emph{European conference on computer vision}, pages 75--91.
  Springer, 2016.

\bibitem[Price et~al.(2009)Price, Morse, and Cohen]{price2009livecut}
Brian~L Price, Bryan~S Morse, and Scott Cohen.
\newblock Livecut: Learning-based interactive video segmentation by evaluation
  of multiple propagated cues.
\newblock In \emph{2009 IEEE 12th International Conference on Computer Vision},
  pages 779--786. IEEE, 2009.

\bibitem[Qi et~al.(2021)Qi, Gao, Hu, Wang, Liu, Bai, Belongie, Yuille, Torr,
  and Bai]{qi2021occluded}
Jiyang Qi, Yan Gao, Yao Hu, Xinggang Wang, Xiaoyu Liu, Xiang Bai, Serge
  Belongie, Alan Yuille, Philip Torr, and Song Bai.
\newblock Occluded video instance segmentation: A benchmark.
\newblock \emph{arXiv preprint arXiv:2102.01558}, 2021.

\bibitem[Tian et~al.(2020)Tian, Shen, and Chen]{tian2020conditional}
Zhi Tian, Chunhua Shen, and Hao Chen.
\newblock Conditional convolutions for instance segmentation.
\newblock In \emph{Computer Vision--ECCV 2020: 16th European Conference,
  Glasgow, UK, August 23--28, 2020, Proceedings, Part I 16}, pages 282--298.
  Springer, 2020.

\bibitem[Tian et~al.(2021)Tian, Shen, Wang, and Chen]{tian2021boxinst}
Zhi Tian, Chunhua Shen, Xinlong Wang, and Hao Chen.
\newblock Boxinst: High-performance instance segmentation with box annotations.
\newblock In \emph{CVPR}, 2021.

\bibitem[Tulsiani and Malik(2015)]{tulsiani2015viewpoints}
Shubham Tulsiani and Jitendra Malik.
\newblock Viewpoints and keypoints.
\newblock In \emph{Proceedings of the IEEE Conference on Computer Vision and
  Pattern Recognition}, pages 1510--1519, 2015.

\bibitem[Wang et~al.(2022{\natexlab{a}})Wang, Cai, Yang, Swaminathan,
  Vasconcelos, Schiele, and Soatto]{wang2022omni}
Pei Wang, Zhaowei Cai, Hao Yang, Gurumurthy Swaminathan, Nuno Vasconcelos,
  Bernt Schiele, and Stefano Soatto.
\newblock Omni-detr: Omni-supervised object detection with transformers.
\newblock \emph{arXiv preprint arXiv:2203.16089}, 2022{\natexlab{a}}.

\bibitem[Wang et~al.(2020)Wang, Kong, Shen, Jiang, and Li]{wang2020solo}
Xinlong Wang, Tao Kong, Chunhua Shen, Yuning Jiang, and Lei Li.
\newblock Solo: Segmenting objects by locations.
\newblock In \emph{ECCV}, 2020.

\bibitem[Wang et~al.(2022{\natexlab{b}})Wang, Yu, De~Mello, Kautz, Anandkumar,
  Shen, and Alvarez]{wang2022freesolo}
Xinlong Wang, Zhiding Yu, Shalini De~Mello, Jan Kautz, Anima Anandkumar,
  Chunhua Shen, and Jose~M Alvarez.
\newblock Freesolo: Learning to segment objects without annotations.
\newblock In \emph{CVPR}, 2022{\natexlab{b}}.

\bibitem[Wang et~al.(2021)Wang, Xu, Wang, Shen, Cheng, Shen, and
  Xia]{wang2021end}
Yuqing Wang, Zhaoliang Xu, Xinlong Wang, Chunhua Shen, Baoshan Cheng, Hao Shen,
  and Huaxia Xia.
\newblock End-to-end video instance segmentation with transformers.
\newblock In \emph{CVPR}, 2021.

\bibitem[Wu et~al.(2021{\natexlab{a}})Wu, Cao, Song, Wang, Yang, and
  Yuan]{wu2021track}
Jialian Wu, Jiale Cao, Liangchen Song, Yu Wang, Ming Yang, and Junsong Yuan.
\newblock Track to detect and segment: An online multi-object tracker.
\newblock In \emph{Proceedings of the IEEE/CVF Conference on Computer Vision
  and Pattern Recognition}, pages 12352--12361, 2021{\natexlab{a}}.

\bibitem[Wu et~al.(2021{\natexlab{b}})Wu, Jiang, Zhang, Bai, and
  Bai]{wu2021seqformer}
Junfeng Wu, Yi Jiang, Wenqing Zhang, Xiang Bai, and Song Bai.
\newblock Seqformer: a frustratingly simple model for video instance
  segmentation.
\newblock \emph{arXiv preprint arXiv:2112.08275}, 2021{\natexlab{b}}.

\bibitem[Wu et~al.(2022)Wu, Liu, Jiang, Bai, Yuille, and Bai]{IDOL}
Junfeng Wu, Qihao Liu, Yi Jiang, Song Bai, Alan Yuille, and Xiang Bai.
\newblock In defense of online models for video instance segmentation.
\newblock In \emph{ECCV}, 2022.

\bibitem[Xie et~al.(2020)Xie, Sun, Song, Wang, Liu, Liang, Shen, and
  Luo]{xie2020polarmask}
Enze Xie, Peize Sun, Xiaoge Song, Wenhai Wang, Xuebo Liu, Ding Liang, Chunhua
  Shen, and Ping Luo.
\newblock Polarmask: Single shot instance segmentation with polar
  representation.
\newblock In \emph{Proceedings of the IEEE/CVF conference on computer vision
  and pattern recognition}, pages 12193--12202, 2020.

\bibitem[Yang et~al.(2019)Yang, Fan, and Xu]{yang2019video}
Linjie Yang, Yuchen Fan, and Ning Xu.
\newblock Video instance segmentation.
\newblock In \emph{ICCV}, 2019.

\bibitem[Yang et~al.(2021{\natexlab{a}})Yang, Fang, Wang, Li, Fang, Shan, Feng,
  and Liu]{yang2021crossover}
Shusheng Yang, Yuxin Fang, Xinggang Wang, Yu Li, Chen Fang, Ying Shan, Bin
  Feng, and Wenyu Liu.
\newblock Crossover learning for fast online video instance segmentation.
\newblock In \emph{ICCV}, 2021{\natexlab{a}}.

\bibitem[Yang et~al.(2021{\natexlab{b}})Yang, Fang, Wang, Li, Shan, Feng, and
  Liu]{yang2021tracking}
Shusheng Yang, Yuxin Fang, Xinggang Wang, Yu Li, Ying Shan, Bin Feng, and Wenyu
  Liu.
\newblock Tracking instances as queries.
\newblock \emph{arXiv preprint arXiv:2106.11963}, 2021{\natexlab{b}}.

\bibitem[Yang et~al.(2022)Yang, Wang, Li, Fang, Fang, Liu, Zhao, and
  Shan]{yang2022tevit}
Shusheng Yang, Xinggang Wang, Yu Li, Yuxin Fang, Jiemin Fang, Liu, Xun Zhao,
  and Ying Shan.
\newblock Temporally efficient vision transformer for video instance
  segmentation.
\newblock In \emph{CVPR}, 2022.

\bibitem[Yu et~al.(2022)Yu, Chen, Wu, Hassan, Li, Yan, Shi, Ye, and
  Han]{yu2022object}
Xuehui Yu, Pengfei Chen, Di Wu, Najmul Hassan, Guorong Li, Junchi Yan, Humphrey
  Shi, Qixiang Ye, and Zhenjun Han.
\newblock Object localization under single coarse point supervision.
\newblock \emph{arXiv preprint arXiv:2203.09338}, 2022.

\bibitem[Zhang et~al.(2020)Zhang, Tian, Shen, You, and Yan]{zhang2020mask}
Rufeng Zhang, Zhi Tian, Chunhua Shen, Mingyu You, and Youliang Yan.
\newblock Mask encoding for single shot instance segmentation.
\newblock In \emph{Proceedings of the IEEE/CVF Conference on Computer Vision
  and Pattern Recognition}, pages 10226--10235, 2020.

\bibitem[Zhou et~al.(2021)Zhou, Li, Li, and Shao]{zhou2021target}
Tianfei Zhou, Jianwu Li, Xueyi Li, and Ling Shao.
\newblock Target-aware object discovery and association for unsupervised video
  multi-object segmentation.
\newblock In \emph{Proceedings of the IEEE/CVF Conference on Computer Vision
  and Pattern Recognition}, pages 6985--6994, 2021.

\bibitem[Zhou et~al.(2018)Zhou, Zhu, Ye, Qiu, and Jiao]{zhou2018weakly}
Yanzhao Zhou, Yi Zhu, Qixiang Ye, Qiang Qiu, and Jianbin Jiao.
\newblock Weakly supervised instance segmentation using class peak response.
\newblock In \emph{Proceedings of the IEEE Conference on Computer Vision and
  Pattern Recognition}, pages 3791--3800, 2018.

\bibitem[Zuffi et~al.(2019)Zuffi, Kanazawa, Berger-Wolf, and
  Black]{zuffi2019three}
Silvia Zuffi, Angjoo Kanazawa, Tanya Berger-Wolf, and Michael~J Black.
\newblock Three-d safari: Learning to estimate zebra pose, shape, and texture
  from images" in the wild".
\newblock In \emph{Proceedings of the IEEE/CVF International Conference on
  Computer Vision}, pages 5359--5368, 2019.

\end{thebibliography}
}


\end{document}